\documentclass[10pt,twocolumn,letterpaper]{article}

\usepackage{iccv}
\usepackage{times}
\usepackage{epsfig}
\usepackage{graphicx}
\usepackage{amsmath}
\usepackage{amssymb}
\usepackage{algorithm}
\usepackage{algorithmic}
\usepackage{multirow}

\usepackage[pagebackref=true,breaklinks=true,letterpaper=true,colorlinks,bookmarks=false]{hyperref}

\iccvfinalcopy % *** Uncomment this line for the final submission

\def\iccvPaperID{389} % *** Enter the ICCV Paper ID here

\ificcvfinal\fi
\begin{document}

%%%%%%%%% TITLE
\title{Learning Complexity-Aware Cascades for Deep Pedestrian Detection}

\author{Zhaowei Cai\\
UCSD\\
{\tt\small zwcai@ucsd.edu}
\and
Mohammad Saberian\\
Yahoo Labs\\
{\tt\small saberian@yahoo-inc.com}
\and
Nuno Vasconcelos\\
UCSD\\
{\tt\small nuno@ucsd.edu}
}

\maketitle
%\thispagestyle{empty}

%%%%%%%%% ABSTRACT
\begin{abstract}
The design of complexity-aware cascaded detectors,
combining
features of very different complexities, is considered. A new cascade design
procedure is introduced,
%based on the minimization of classification
%risk under a complexity constraint.
%The latter is
%defined with resort to a new complexity risk, which combines a
%monotonic loss function and a complexity measure akin to the
%classification margin.
by %This forulates
formulating cascade learning as the Lagrangian
optimization of a risk that accounts for both accuracy and complexity.
A boosting algorithm, denoted as {\it complexity aware cascade training}
(CompACT),  is then derived to solve this optimization. CompACT cascades
are shown to seek an optimal trade-off between accuracy and complexity by
pushing features of higher complexity to the later cascade stages, where only
a few difficult candidate patches remain to be classified. This
enables the use of features of vastly different complexities in a
single detector. In result, the feature pool can be expanded to
features previously impractical for cascade
design, such as the responses of a deep convolutional neural network (CNN).
This is demonstrated through the design of a pedestrian detector
with a pool of features whose complexities span orders of magnitude.
The resulting cascade generalizes the combination of a CNN with
an object proposal mechanism: rather than a pre-processing stage, CompACT
cascades seamlessly integrate CNNs in their stages. This enables
state of the art performance on the Caltech and KITTI datasets, at fairly
fast speeds.
\end{abstract}

%============================================================================================
\section{Introduction}
\label{sec:intro}

Pedestrian detection is an important problem in computer vision. Many of
its applications, e.g. smart vehicles or surveillance, require real-time
detection. Since, under the popular sliding window paradigm, there are
close to a million windows per 640$\times$480 pixel image, detection
complexity can easily become intractable. This is an impediment to the
deployment of sophisticated classifiers, such as deep learning models,
in the pedestrian detection arena. The most popular architecture for
real-time object detection is the detector cascade
of~\cite{DBLP:journals/ijcv/ViolaJ04}. It exploits the fact that most
image patches can be assigned to the background class by evaluation
of a few simple cascade stages. This guarantees computational efficiency
without compromising accuracy, since the few resulting false positives can be
rejected by more complex detectors, in the late cascade stages. Given that
these are rarely used, their complexity is not an impediment to
high detection speeds. In result, it is possible to have both efficient
and accurate detection.

While the cascade detection principle is intuitive, its implementation
is far from trivial. Early cascade designs required extensive heuristics
to determine the cascade
configuration~\cite{DBLP:journals/ijcv/ViolaJ04,DBLP:conf/iccv/XiaoZZ03,
DBLP:conf/cvpr/BourdevB05}, lacking the ability to explicitly optimize the
trade-off between detection accuracy and complexity. A commonly used
assumption is that all features have equivalent complexity. This
significantly simplifies the design, which reduces to choosing the features
that maximize detection accuracy. In fact, popular
methods~\cite{DBLP:conf/cvpr/BourdevB05,
DBLP:journals/pami/DollarABP14} simply use a boosting algorithm
(typically AdaBoost~\cite{DBLP:conf/eurocolt/FreundS95}) to design a
non-cascaded classifier and then transform it into a cascade, by addition
of thresholds. These approaches suffer from two main
problems. First, they do not aim to select features that optimize
the trade-off between detection accuracy and complexity. Second,
the ``equivalent feature complexity'' hypothesis only produces sensible
cascades when applied to features that indeed have similar complexity.
This constraint is, however, frequently
violated \cite{DBLP:conf/cvpr/BenensonMTG13,
DBLP:conf/eccv/PaisitkriangkraiSH14,DBLP:journals/corr/ZhangBS15}.

In fact, it has been remarkably difficult to accommodate, in cascade
learning, features significantly heavier than those in common use.
This problem is particularly pressing given the recent success of deep
learning in object
recognition~\cite{DBLP:conf/nips/KrizhevskySH12,DBLP:journals/corr/SimonyanZ14a}.
The intractable computation of a deep learning model under the sliding
window paradigm is usually addressed with recourse to object proposal
mechanisms~\cite{DBLP:conf/iccv/SandeUGS11}, giving rise to a two-stage
cascade that is far from optimal, in terms of the trade-off between detection
accuracy and speed. For pedestrian detection, object proposals are
frequently implemented with weak pedestrian detectors, sometimes cascaded
detectors themselves~\cite{DBLP:journals/corr/HosangOBS15}. Due to the ad-hoc
nature of these solutions, deep learning models have not been competitive
for pedestrian detection, contradicting their recognition and
classification performance~\cite{DBLP:conf/nips/KrizhevskySH12,
DBLP:journals/corr/SimonyanZ14a}.

In this work, we address these problems by seeking an algorithm for optimal
cascade learning under a criterion that penalizes {\it both\/} detection errors
and complexity. For the latter, we introduce a  measure of {\it implementation
complexity\/} that allows the definition of a {\it complexity risk\/}
akin to the empirical risk commonly used for classifier design.
This makes it possible to define quantities such as complexity margins and
complexity losses, and account for these in the learning process. We do this
with recourse to a Lagrangian formulation, which optimizes for the usual
classification risk under a constraint in the complexity risk. A boosting
algorithm that minimizes this Lagrangian is then derived. This algorithm,
denoted {\it Complexity-Aware Cascade Training\/} (CompACT), is shown to
select inexpensive features in the early cascade stages, pushing the
more expensive ones to the later stages. This enables the
combination of features of vastly different complexities in a single detector.
% The resulting CompACT cascades can be seen as a generalization of the
% object proposal architecture, which seamlessly integrates CNNs within
% detection stages of many other levels of complexity.
These properties are demonstrated by the successful application of CompACT to
the problem of pedestrian detection, using a pool of features
ranging from Haar wavelets to deep convolutional neural networks (CNNs).
%, which provides a good coverage of the discrimination vs complexity spectrum.

Overall, this work makes three major contributions. First, it proposes a
novel algorithm for learning a complexity aware cascade, so as to achieve
an optimal trade-off between accuracy and speed. To the best of our
knowledge, this is the first algorithm to {\it explicitly\/} account for
variable feature complexity in cascade learning, supporting weak
learners of widely different complexities. Second, CompACT seamlessly
integrates handcrafted and CNN features in a unified detector.
This generalizes the object proposal architecture, guaranteeing
the seamlessly integration of CNN stages with stages of
any other complexity. Finally, a ComPACT cascade for pedestrian detection
is shown to achieve state of the art results on both
Caltech \cite{DBLP:journals/pami/DollarWSP12} and
KITTI \cite{DBLP:conf/cvpr/GeigerLU12}, at faster speeds than
the closest competitors.

%============================================================================================
\section{Related Works}

Detector cascades learned with boosting are commonly used for detecting
template-like objects, e.g.
faces \cite{DBLP:journals/ijcv/ViolaJ04,DBLP:conf/cvpr/BourdevB05,
DBLP:conf/iccv/XiaoZZ03,DBLP:conf/iccv/XiaoZST07},
pedestrians \cite{DBLP:journals/pami/DollarABP14,
DBLP:journals/pami/SaberianV12}, or
cars \cite{DBLP:journals/jmlr/SaberianV14}. Early
approaches used heuristics to find a cascade configuration of
good trade-off between classification accuracy and
complexity \cite{DBLP:journals/ijcv/ViolaJ04,DBLP:conf/cvpr/BourdevB05,
DBLP:conf/iccv/XiaoZZ03,DBLP:conf/iccv/XiaoZST07}. More recently,
optimization of the accuracy-complexity trade-off has started to
receive attention~\cite{DBLP:conf/iccv/Masnadi-ShiraziV07,
DBLP:journals/pami/SaberianV12,DBLP:journals/jmlr/SaberianV14,
DBLP:conf/cvpr/ZhengL09}.
\cite{DBLP:conf/cvpr/ZhengL09} empirically added a complexity
term to the objective function of RealBoost.
\cite{DBLP:conf/iccv/Masnadi-ShiraziV07,
DBLP:journals/pami/SaberianV12,DBLP:journals/jmlr/SaberianV14} introduced the Lagrangian formulation that we
adopt, but use a single feature family throughout the cascade.
Since early cascades stages must be very efficient, this implies adopting
simple weak learners, e.g. decision stumps.
%over fast features \cite{DBLP:journals/ijcv/ViolaJ04,
%DBLP:journals/pami/DollarABP14}.

% However, this  makes it difficult to
% to design accurate detectors in the later stages of the cascade.

This has motivated extensive work on the design of efficient features.
For pedestrian detection, the integral channel features
of \cite{DBLP:conf/bmvc/DollarTPB09} have recently become popular.
They extend the Haar-like features of \cite{DBLP:journals/ijcv/ViolaJ04}
into a set of color and histogram-of-gradients (HOG) channels. More
recently, a computationally
efficient version of \cite{DBLP:journals/ijcv/ViolaJ04}, denoted the
aggregate channel features (ACF), has been introduced
in \cite{DBLP:journals/pami/DollarABP14}.
\cite{DBLP:conf/eccv/PaisitkriangkraiSH14} complemented ACF with local
binary patterns (LBP) and covariance features, for better detection accuracy.

Several works proposed alternative feature channels, obtained
by convolving different filters with the original HOG+LUV channels  \cite{DBLP:conf/cvpr/ZhangBC14,DBLP:journals/corr/ZhangBS15,
DBLP:conf/cvpr/BenensonMTG13,DBLP:conf/nips/NamDH14}.
The SquaresChnFtrs of \cite{DBLP:conf/cvpr/BenensonMTG13} reduce the
large number of features of
\cite{DBLP:conf/bmvc/DollarTPB09,DBLP:journals/ijcv/ViolaJ04} to
16 box-like filters of various sizes.
\cite{DBLP:conf/nips/NamDH14} extended the locally decorrelated
features of \cite{DBLP:conf/eccv/HariharanMR12} to ACF,
learning four 5$\times$5 PCA-like filters from each of the ACF
channels. Instead of empirical filter design, Zhang et
al \cite{DBLP:conf/cvpr/ZhangBC14} exploited prior knowledge
about pedestrian shape to design informed filters. They later found, however,
that such filters are actually not
needed \cite{DBLP:journals/corr/ZhangBS15}. Instead, the number of filters
appears to be the most important variable: features as simple as
checkerboard-like patterns, or purely random filters, can achieve very
good performance, as long as there are enough of them. Although reached state-of-the-art
performance has been achieved \cite{DBLP:conf/eccv/PaisitkriangkraiSH14,
DBLP:journals/corr/ZhangBS15}, they are relatively slow, due to the convolution computations with several hundred filters.

While deep convolutional learning classifiers have achieved impressive results for general object detection
\cite{DBLP:conf/cvpr/GirshickDDM14,DBLP:conf/eccv/HeZR014}, e.g.
on VOC2007 or ImageNet, they have not excelled on pedestrian
detection \cite{DBLP:conf/cvpr/SermanetKCL13,DBLP:conf/iccv/OuyangW13}. Benchmarks like Caltech
\cite{DBLP:journals/pami/DollarWSP12} are still dominated by
classical handcrafted features (see e.g. a recent comprehensive evaluation of
pedestrian detectors by \cite{Benenson2014Eccvw}). Recently,
\cite{DBLP:journals/corr/HosangOBS15} transferred the R-CNN
framework to the pedestrian detection task, showing some improvement over
previous deep learning detectors \cite{DBLP:conf/cvpr/SermanetKCL13,DBLP:conf/iccv/OuyangW13}. However, the gap to the state of the art is still
significant. Deep models also tend to be too heavy for sliding window
detection. This is usually addressed with object proposal
mechanisms \cite{DBLP:conf/cvpr/GirshickDDM14,DBLP:conf/iccv/WangYZL13,
DBLP:journals/corr/HosangOBS15} that pre-select the most promising image
patches. This two-stage decomposition (proposal generation and
classification) is a simple cascade mechanism. In this work, we consider the
seamless combination of these two stages into a cascade explicitly
designed to account for both accuracy and complexity, so as to achieve
detectors that are both highly accurate and fast.

%============================================================================================
\section{Complexity-Aware Cascade Training}
\label{sec:complextiy-aware cascade}

In this section we introduce the CompACT algorithm.

%-----------------------------------------------------------------------------------
\subsection{AdaBoost}

A decision rule $h(x) = sign[F(x)]$ of predictor $F(x)$ maps a
feature vector $x \in {\cal X}$ to a class label $y \in {\cal Y} = \{-1,1\}$.
Boosting learns a strong decision rule by combining a set of weaker
learners $f_k(x)$,
\begin{equation}
  F(x)=\sum_{k}f_{k}(x),
  \label{eq:additive}
\end{equation}
using functional gradient descent on a classification risk
\cite{DBLP:journals/AS/Friedman00,DBLP:conf/nips/MasonBBF99}.
AdaBoost \cite{DBLP:conf/eurocolt/FreundS95} is based on the exponential
loss $\phi(yF(x)) = e^{-yF(x)},$ minimizing %the risk
%\begin{equation}
%\label{equ:adaboost loss}
%\mathcal{R}_{E}[F]=E_{X,Y}[e^{-yF(x)}].
%\end{equation}
%When learning from a finite training sample $\{(x_i,y_i)\}$ this is
%approximated by
the empirical risk
\begin{equation}
\mathcal{R}_{E}[F]\simeq\frac{1}{|S_{t}|}\sum_{i}e^{-y_{i}F(x_{i})},
\label{equ:adaboost emprisk}
\end{equation}
on training samples $S_t=\{(x_i,y_i)\}$.
Boosting iterations compute the functional derivative of (\ref{equ:adaboost emprisk}) along the direction of weak learner $g(x)$ at the location of the
current predictor $F(x)$,
\begin{align}
\label{equ:risk derivative}
<\delta{\mathcal{R}_{E}}[F],g>&=\frac{d}{d\epsilon}\mathcal{R}_{E}[F+\epsilon{g}]\big|_{\epsilon=0}\nonumber\\
&=\frac{1}{|S_{t}|}\sum_{i}\Big[\frac{d}{d\epsilon}e^{-y_{i}(F(x_{i})+\epsilon{g}(x_{i}))}\Big]\Big|_{\epsilon=0}\nonumber\\
&=-\frac{1}{|S_{t}|}\sum_{i}y_{i}w_{i}g(x_{i}),
\end{align}
where
\begin{equation}
w_{i}=w(x_{i})=e^{-y_{i}F(x_{i})}.
\end{equation}
The predictor is updated by selecting the steepest descent direction
within a weak learner pool $\textbf{G}=\{g_{1}(x),\cdots, g_{n}(x)\}$,
\begin{align}
\label{equ:best risk wl}
g^{*}(x)&=\arg\max_{g\in{\textbf{G}}}<-\delta{\mathcal{R}_{E}}[F],g>\nonumber\\
&=\arg\max_{g\in{\textbf{G}}}\frac{1}{|S_{t}|}\sum_{i}y_{i}w_{i}g(x_{i}).
\end{align}
The optimal step size for the update is
\begin{equation}
\alpha^{*}=\arg\min_{\alpha}\mathcal{R}_{E}[F+\alpha{g^{*}}].
\end{equation}
For binary $g^{*}(x)$, this has a closed form solution
\begin{equation}
\label{equ:closed form alpha}
\alpha^{*}=\frac{1}{2}\log\frac{\sum_{i|y_{i}=g^{*}(x)}w_{i}^{k}}
{\sum_{i|y_{i}\neq{g}^{*}(x)}w_{i}^{k}}.
\end{equation}
Otherwise, the optimal step size is found by a line search.

%---------------------------------------------------------------------------------
\subsection{Complexity-Aware Learning}

Complexity-aware learning aims for the best trade-off
between classification accuracy and complexity. This can be
formulated as a constrained optimization problem,
where classification risk is minimized under a bound on a complexity
risk $R_C[F]$,
\begin{equation}
F^*(x) = \arg\min_F R_E[F] \quad  s.t. \quad R_C[F] < \gamma,
\end{equation}
and is identical to the minimization of the Lagrangian
\begin{equation}
\label{equ:lagrangian}
\mathcal{L}[F]=\mathcal{R}_{E}[F]+\eta\mathcal{R}_{C}[F],
\end{equation}
where $\eta$ is a Lagrange multiplier that only depends on $\gamma$.
To define a complexity risk, we note that~(\ref{equ:adaboost emprisk})
can be written as
\begin{equation}
\mathcal{R}_{E}[F]\simeq\frac{1}{|S_{t}|}\sum_{i}\phi[\xi(y_i,F(x_{i}))],
\label{eq:emprisk}
\end{equation}
with $\phi(v)=e^{-v}$ and $\xi(y,F(x))=yF(x)$. The function $\xi(.)$
is the margin of example $x$ under predictor $F(.)$ and measures
the confidence of the classification. Large positive margins indicate
that $x$ is correctly classified with high confidence, large negative
margins the same for incorrect classification, and a margin zero that
the example is on the classification boundary. The loss
$\phi(.)$ is usually monotonically decreasing, penalizing all examples with
less than a small positive margin. This forces the learning algorithm
to concentrate on these examples, so as to produce as few negative margins
as possible. The exponential loss of AdaBoost makes the penalty exponential
on the confidence of incorrectly classified examples.

In this work, we consider complexity risks of a similar form
\begin{equation}
\mathcal{R}_{C}[F]\simeq\frac{1}{|S_{t}|}\sum_{i}\tau[\kappa(y_i,F(x_{i}))],
\label{eq:compelxity emprisk}
\end{equation}
where $\kappa[y,F(x)]$ is a measure of complexity for the classification
of example $x$ under $F(.)$ and $\tau(.)$ a non-negative
loss function
that penalizes complexity. Drawing inspiration from the classification
risk, we measure complexity with the complexity margin
\begin{equation}
\kappa[y,F(x)] = y \Omega(F(x)),
\label{eq:kappa}
\end{equation}
where $\Omega(F(x))$ is a function of the time required to
evaluate $F(x)$, e.g. a number of machine operations or some other
empirical measure of complexity. The complexity margin
of~(\ref{eq:kappa}) assigns positive (negative)
complexity to positive (negative) examples, reflecting the fact that
the computation spent on negative examples is ``wasted'' or
``negative'' while that spent on positives is ``justified'' or ``positive''.
%reflecting the fact that
%positives are expected to require more complexity than negatives.
While positives have to survive all cascade stages, negatives
should be rejected with little computation.
%In this sense,
The complexity loss $\tau(v)$ then determines the complexity-aware behavior
of learning algorithms. For example, a decreasing
$\tau(v)$  for $v<0$, penalizes negative
examples of large complexity. This encourages classifiers that reject
negatives with as little computation as possible.
On the other
hand, an increasing $\tau(v)$ for $v>0$ penalizes positives
of large complexity.

%---------------------------------------------------------------------------------
\subsection{Embedded Cascade}

A cascaded classifier is implemented as a sequence of classification
stages $h_i(x) = sgn[F_i(x)+T_{i}]$, where $T_{i}$ is a threshold. A
popular architecture is the embedded cascade, whose predictor has the embedded structure,
\begin{equation}
  F_{k}(x) = F_{k-1}(x)+ f_{k}(x) = \sum_{j=1}^{k} f_j(x).
\end{equation}
In this paper, the cascade complexity is measured by the average per stage complexity,
\begin{equation}
\label{equ:average complexity}
\Omega(F(x))=\frac{1}{m}\sum_{k=1}^{m}r_{k}(x)\Omega{(f_{k}(x))},
\end{equation}
where, using $u[\cdot]$ to denote the Heaviside step function,
\begin{equation}
\label{eq:rejection}
r_{k}(x)=\prod_{j=1}^{k-1}u\big[F_{j}(x)+T_{j}\big],
\end{equation}
is an indicator of examples that survive all stages
prior to $k$, i.e. $r_{k}(x) = 1$ if $F_i(x)+T_i>0, \forall i<k$, and
$r_k(x) = 0$ otherwise. Since the average complexity is bounded
by the largest weak learner complexity, it leads to a more balanced Lagrangian
in (\ref{equ:lagrangian}) than the total complexity.

%----------------------------------------------------------------------------
\subsection{Cascade Boosting}

The minimization of~(\ref{equ:lagrangian}) requires the
functional derivative of the Lagrangian along the direction of weak
learner $g(x)$ at the location of the current predictor $F(x)$,
\begin{equation}
  \label{equ:lagrangianD}
  <\delta \mathcal{L}[F], g>
  =<\delta\mathcal{R}_{E}[F],g>+\eta<\delta\mathcal{R}_{C}[F],g>,
\end{equation}
where $<\delta\mathcal{R}_{E}[F],g>$ is as in (\ref{equ:risk derivative}).
To compute the derivative of the complexity risk we define $u(\epsilon)$ as $u(\epsilon) = 1$ for $\epsilon >0$ and $u(\epsilon) = 0$ otherwise, and write
\begin{eqnarray*}
  \lefteqn{\Omega(F(x) + \epsilon g(x)) =}
  && \\
  &=& \Omega(F(x)) +
      u(\epsilon)\big[\Omega(F(x)+g(x))-\Omega(F(x))\big]\nonumber\\
  &=& \Omega(F(x)) [1-u(\epsilon)]
      +u(\epsilon)\Omega(F(x)+g(x))\nonumber\\
  &=& \Omega(F(x)) [1-u(\epsilon)] \\
  &+&  \frac{u(\epsilon)}{m+1}
      \big[\sum_{k=1}^{m}r_{k}(x)\Omega{(f_{k}(x))} + r_{m+1}(x) \Omega{(g(x))}
      \big] \nonumber \\
  &=& \Omega(F(x)) \left[1-u(\epsilon) + \frac{m}{m+1}u(\epsilon)\right]
      \nonumber \\
  &+&  \frac{u(\epsilon)}{m+1} r_{m+1}(x) \Omega{(g(x))} \nonumber \\
  &=&\Omega(F(x))[1-u(\epsilon)\zeta_m]+u(\epsilon)\frac{r_{m+1}(x)}{m+1}\Omega(g(x)),
\end{eqnarray*}
%\begin{align}
%&\Omega(F(x) + \epsilon r_{m+1}(x)g(x))\nonumber\\
%&=\Omega(F(x))+u(\epsilon)\big[\Omega(F(x)+r_{m+1}(x)g(x))-\Omega(F(x))\big]\nonumber\\
%&=\Omega(F(x)[1-u(\epsilon)\zeta_m]+u(\epsilon)\frac{r_{m+1}}{m+1}\Omega(g(x))
%\end{align}
where $\zeta_m = 1- \frac{m}{m+1}$ and we have used~(\ref{equ:average complexity}). Since $u(\epsilon)$  is not differentiable, it is approximated by
$u(\epsilon) \approx \sigma(\epsilon)$, where $\sigma(\epsilon)$ is a
differentiable function with $\sigma(0) = 0$, leading to
\begin{align}
\label{equ:complexity derivative}
&<\delta{\mathcal{R}_{C}}[F],g>\\
&=\frac{1}{|S_{t}|}\sum_{i}
\bigg[\frac{d}{d\epsilon}\tau\Big[y_i\Omega\Big(F(x_i)
+\epsilon g(x_i)\Big)\Big]\bigg]\bigg|_{\epsilon=0}
\nonumber\\
=&-\frac{1}{|S_{t}|}\sum_{i}y_i\psi(y_i,x_i)\bigg[\frac{r_{m+1}(x_i)}{m+1}\Omega(g(x_i))-\zeta_m\Omega(F(x_i))\bigg],\nonumber
\end{align}
where
\begin{equation}
\psi(y_i,x_i) = -\tau^\prime\left[y_i \Omega(F(x_i))\right] \sigma^\prime(0).
\label{eq:psi}
\end{equation}

%Using (\ref{equ:lagrangian}), (\ref{equ:risk derivative}) and (\ref{equ:complexity derivative}),
Each boosting iteration updates $F(x)$ with a step along the steepest descent
direction of~(\ref{equ:lagrangianD}) within the weak learner learner pool
$\textbf{G}$,
\begin{align}
\label{eq:bestLag}
g^{*}(x)&=\arg\max_{g\in{\textbf{G}}}<-\delta{\mathcal{L}}[F],g>.
\end{align}
Combining~(\ref{equ:risk derivative}), (\ref{equ:lagrangianD}), and
(\ref{equ:complexity derivative}) and denoting $r_i = r_{m+1}(x_i)$,
$\omega_i = \omega(y_i,x_i)$, $g_i = g(x_i),$ and $\psi_i = \psi(y_i,x_i)$,
this is the direction that maximizes
\begin{equation}
%  <-\delta{\mathcal{L}}[F],g> =
\mathcal{D}[g] =
\frac{1}{|S_{t}|}\sum_{i} y_i \left[
\omega_i g_i + \frac{\eta r_i \psi_i \Omega(g_i)}{m+1} \right].
\label{eq:negGrad}
\end{equation}
Note that the term $\zeta_m\Omega(F(x_i))$ of
(\ref{equ:complexity derivative}) does not depend on $g$ and plays no
role in the optimization.
The optimal step size for the update is
\begin{equation}
\alpha^{*}=\arg\min_{\alpha}\mathcal{L}[F+\alpha{g^{*}}],
\label{eq:a*}
\end{equation}
and can be found by a line search. The cascade predictor is finally
updated with
\begin{equation}
F^{new}(x) = F(x) + \alpha^* g^*(x).
\label{eq:update}
\end{equation}
Note that, from~(\ref{eq:psi}), $\sigma^\prime(0)$  is a constant that rescales all $\psi_i$ equally. Hence, in~(\ref{eq:negGrad}), it can be absorbed into $\eta$. Without loss of generality, we assume that $\sigma^\prime(0) = 1$. This boosting algorithm is denoted the {\it complexity aware cascade training\/} (CompACT) boosting algorithm.

%%%%%%%%%%%%%%%%%%%%%%%%%%%%%%%%%%%%% algorithm %%%%%%%%%%%%%%%%%%%%%%%%%%%%%%%%%%%%%%

%\begin{algorithm}[t]
%\caption{\small{Complexity-Aware Cascade Learning}} \label{alg:cascade learning}
%{\small
%\begin{algorithmic}
%\STATE \textbf{Input:} \STATE Training images set $\mathcal{I}$; positive training samples $S_{t}^{+}$; the number of weak learners $M$; Lagrangian multiplier $\eta$; weak learner pool $\textbf{G}$; the false positive rate threshold $\theta_{fp}$.
%\STATE \textbf{Initialization:} \STATE $\mathcal{F}=F_{0}=0$; randomly sample negative instances $S_{t}^{-}$ from $\mathcal{I}$.
%\STATE \textbf{Output:} \STATE The cascade $\mathcal{F}=\{F_{1},F_{2},\cdots,F_{M}\}$.
%\WHILE{$k<M$}
%\STATE 1. Select the best weak learner $g^{*}$ with (\ref{eq:bestLag}) from the pool $\textbf{G}$, where $<-\delta{\mathcal{L}}[F],g>$ is gotten from (\ref{equ:final lagragian}), and find the optimal step size $\alpha^{*}$ with (\ref{eq:a*}).
%\STATE 2. Generate the $k^{th}$ cascade stage $F_{k}=F_{k-1}+\alpha^{*}g^{*}$.
%\STATE 3. Stack the new stage into the cascade $\mathcal{F}=\mathcal{F}\cup{F_{k}}$
%\STATE 4. Perform current cascade $\mathcal{F}$ on current negative training set $S_{t}^{-}$ and get the false positive rate $fp$.
%\IF{$fp<\theta_{fp}$}
%\STATE 5. Bootstrap the new $S_{t}^{-}$ from the image set $\mathcal{I}$.
%\ENDIF
%\STATE 6. $k=k+1$
%\ENDWHILE
%\end{algorithmic}}
%\end{algorithm}

%-----------------------------------------------------------------------------
\subsection{Properties}

CompACT has a number of interesting properties. First, the contribution of each
training example to the complexity term in (\ref{eq:negGrad}) is multiplied by $r_i$. Hence, only examples that survive the current cascade $F$ contribute to
the complexity term. We refer to the $x_i$ such that $r_i = 1$ as
{\it active\/} examples.  Note that, given the set of active examples
\begin{equation}
S_a(F) =\{ (x_i,y_i) \in S_t | r_i = 1\},
\end{equation}
associated with $F$, (\ref{eq:negGrad}) can be replaced by
\begin{equation}
%  <-\delta{\mathcal{L}}[F],g>_a =
  \mathcal{D}[g] =
  \frac{1}{|S_t|} \left( \sum_{i} y_i \omega_i g_i
    + \sum_{i|r_i = 1} y_i \frac{\eta \psi_i \Omega(g_i)}{m+1} \right).
 \label{eq:negGrad2}
\end{equation}
This complies with the intuition that examples which do not reach stage $m+1$ during the cascade operation should not affect the complexity term for that stage.

Second, most implementations of cascaded classifiers use weak learners of
example-independent complexity, i.e. $\Omega(g(x_i)) = \Omega_g, \forall i$.
While this does not hold for the cascade in general (different examples can
be rejected at different stages), it holds for the examples in $S_a$,
i.e. $\Omega(F(x_i)) = \Omega_F, \forall x_i \in S_a$. In this case, the
complexity weights only depend on the label $y_i$. Defining
$\psi^+=-\tau^\prime[\Omega_F]$ ($\psi^- = -\tau^\prime[-\Omega_F]$)
as the value of $\psi_i$ for positive (negative) examples, and
$\pi_F^-$ ($\pi_F^+$) as the percentage of negative (positive) active
examples, (\ref{eq:negGrad}) reduces to
%\begin{align}
\begin{equation}
\mathcal{D}[g] = \frac{1}{|S_t|}\sum_{i} y_i \omega_{i}g(x_i)\\
-\frac{\eta}{m+1} \frac{|S_a|}{|S_t|}\xi_F \Omega_{g},
\label{equ:final lagragian}
\end{equation}
with $\xi_F  = \pi^{-}_{F}\psi_{F}^{-}-\pi^{+}_{F}\psi_{F}^{+}$.
%
%  \mathcal{D}[g] = &\frac{1}{|S_a(F)|}\sum_{i} y_i \omega_{i}g(x_i)\\
%  &-\frac{\eta}{m+1}(\pi^{-}_{F}\psi_{F}^{-}-\pi^{+}_{F}\psi_{F}^{+})\Omega_{g}\nonumber
%
%\end{align}
Since $|S_a|$ decreases with cascade length, the rescaling of
$\eta$ by $\frac{|S_a|}{|S_t|}$ gradually weakens the complexity
constraint as the cascade grows. While in the early iterations there
is pressure to select weak learners of reduced complexity,
this pressure reduces as iterations progress. Gradually, complex weak learners are penalized less and the algorithm asymptotically reduces to a cascaded version of AdaBoost.
This makes intuitive sense, since the latter
cascade stages process a much smaller percentage of the examples
than the earlier ones and have much less impact on the overall complexity.
On the other hand, since the surviving examples are the most difficult to
classify, accurate classification requires weak learner
accuracy to increase  with cascade length. This usually (but not always) implies that weak learner complexity increases as well because powerful features usually require heavy computation. By pushing the complexity to the later
stages, the algorithm can learn cascades that are {\it both\/} accurate and
computationally efficient. This effect is reinforced by the fact
that $1/(m+1)$ also decreases with cascade length.

The loss $\tau(v)$ enables fine-tuning of this general behavior,
via $\xi_F$. In this work, we adopt the hinge loss
$\tau(v)=\max(0,-v)$, for which $\psi_{F}^{-}=1, \psi_{F}^{+}=0$ and
$\xi_F = \pi_F^-$. This assigns no penalty to the complexity of positive
examples, encouraging CompACT to focus on the fast rejection of negatives.

%============================================================================================
\section{Pedestrian Detection}

This section discusses the proposed pedestrian detector.

%-------------------------------------------------------------------------
\subsection{Feature Pools of Variable Complexity}
\label{sec:features}

%Most works in cascaded detector design rely on the feature
%set that leads to the best detection performance. This is
%sensible under the standard cascade design paradigm, where
%all features have identical complexity. In this case, there
%is no benefit in choosing any features other than those that
%maximize detection accuracy. On the other hand,
CompACT seeks the optimal trade-off between accuracy and
complexity, at each cascade stage. This is most effective when
the feature pool is composed of features of various complexities.
In the cascade literature, where most detectors use a single feature family,
it is common practice to pre-compute a large number of feature responses
at all image locations, before any detection takes place
\cite{DBLP:conf/nips/NamDH14,DBLP:journals/corr/ZhangBS15,
DBLP:conf/eccv/PaisitkriangkraiSH14}. This, however, has unfeasible complexity
if the feature pool is very large (e.g. the 200,000$\thicksim$500,000
features proposed per patch in \cite{DBLP:journals/corr/ZhangBS15,
DBLP:conf/eccv/PaisitkriangkraiSH14}) or some features are
computationally intense (e.g. the CNN features
of \cite{DBLP:conf/nips/KrizhevskySH12,DBLP:journals/corr/SimonyanZ14a}).
In these cases, it is neither tractable nor necessary to pre-compute all
features at all image locations. For example, a cascade of 2048 decision
trees of depth 2, will evaluate at most 4096 features per patch. Since the
cascade rejects most candidate patches after a few stages, the
most intensive features (e.g. CNN) are unlikely to be needed at most
image locations. Hence, while pre-computation is useful for
low-complexity features, complex features should be evaluated as necessary.
We refer to the former as {\it pre-computed\/} features and the latter
as {\it computed just-in-time\/} (JIT).

\subsubsection{Pre-computed Features}

Our pre-computed feature set consists of
ACF \cite{DBLP:journals/pami/DollarABP14}, mostly due to its
computational efficiency. Following \cite{DBLP:journals/pami/DollarABP14}, we
extract 10 LUV+HOG channels. Since these are pre-computed, the
complexity of using an ACF feature in any cascade stage is 1.

\subsubsection{Just-in-time Features}

The JIT pool contains several feature subsets. The ability to weigh accuracy vs. computation enables CompACT to seamlessly combine these feature sets.

\vspace{.05in}
\noindent{\bf SS:}
The self-similarity (SS) features of \cite{DBLP:conf/cvpr/ShechtmanI07} capture
the difference between local patches and have achieved good performance on
edge detection tasks \cite{DBLP:conf/cvpr/LimZD13,DBLP:conf/iccv/DollarZ13}.
Following \cite{DBLP:conf/cvpr/LimZD13,DBLP:conf/iccv/DollarZ13}, we compute
SS features on a 12$\times$6 grid of the 16$\times$8 ACF patch.
This results in $\begin{pmatrix}72\\2\end{pmatrix}\times{10}=25,560$ SS
features per patch. Since the computation of an SS feature involves 2
ACF values, its complexity is 2.

%%%%%%%%%%%%%%%%%%%%%%%%%%%% checkerboard %%%%%%%%%%%%%%%%%%%%%%%%%%%
\begin{figure}[!t]
\centering
\centerline{\epsfig{figure=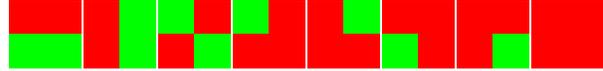,width=8cm,height=1cm}}
\caption{Eight 2$\times$2 checkerboard-like filters used in this work.
  Red (Green) is used to represent value +1 (-1).}
\label{fig:checkerboard}
\end{figure}

\vspace{.05in}
\noindent{\bf CB:}
Checkerboard features (CB) are the result of convolving the ACF channels
with a set of checkerboard filters. \cite{DBLP:journals/corr/ZhangBS15}
has shown that a simple set of such features could achieve state-of-the-art
performance for pedestrian detection. Based on their observation that the
number of features determines performance (rather than feature type), we
adopt the set of 8 simple 2$\times$2 checkerboard filters of
Figure \ref{fig:checkerboard}. A CB has implementation complexity of 4.

\vspace{.05in}
\noindent{\bf LDA:}
Locally decorrelated HOG features, computed with linear discriminant
analysis (LDA), have shown some superiority for object detection over
HOG features \cite{DBLP:conf/eccv/HariharanMR12}.
\cite{DBLP:conf/nips/NamDH14} showed that the computation of these features
on ACF channels leads to a big improvement over ACF. We adopt this feature
family but, unlike \cite{DBLP:conf/nips/NamDH14}, restrict the filter
size to 3$\times$3. LDA features have complexity 9.

\vspace{.05in}
\noindent{\bf CNN:}
In addition to operators defined over the ACF channels, we consider a set
of CNN features. The CNN is a smaller version of the popular model
of \cite{DBLP:conf/nips/KrizhevskySH12}, with five convolutional layers
and one fully connected layer. The CNN is applied to 64$\times$64 image
patches, the first convolutional layer has 32 filters, the remaining
four have 64, and the fully connected layer consists of 1024
hidden units. All convolutional filters have size 3$\times$3, and
stride 1. The CNN model was originally trained with the ILSVRC14-DET
dataset \cite{ILSVRC15}, using the cropped object patches, and then fine
tuned on the target pedestrian dataset. For feature extraction, we only use
the output of the $5^{th}$ convolutional layer, which can be seen as CNN
feature channels, similar to ACF. These features are denoted as CNN.
Inspired by the good performance and simplicity of the
checkerboard features on ACF, we also
compute them on the conv5 feature channels. These are denoted CNNCB features.

The complexity of CNN features is of a different nature than that of
ACF features. First, the implementation on a different
processor (GPU instead of CPU) makes the direct comparison of
number of operations meaningless. Second, while the CNN features
are computed on an ``as needed'' basis, the structure of the
network makes it inefficient to compute each feature individually.
If the CNN features are needed to classify a certain image window,
it is significantly more efficient to compute the $5^{th}$ layer
responses over the whole window than repeatedly applying the
network to sub-window regions. We account for these difficulties
by setting a trigger complexity $\Omega_{CNN}$ for
CNN features. That is, in (\ref{equ:final lagragian}),
CNN features have $\Omega_g = \Omega_{CNN}$ if no CNN feature has been used
by the previous cascade stages to classify the current patch. Once
the CNN features are
computed, the complexity of using any CNN feature is $1$, similar to ACF,
while CNNCB features have complexity $4$.

\subsection{Embedding Large CNN Models}
\label{subsec:big cnn}

Large CNN models \cite{DBLP:conf/nips/KrizhevskySH12,
DBLP:journals/corr/SimonyanZ14a} are now popular in computer vision. However,
the use of these models in CompACT is challenging,
due to the computational cost of embedding them in the iterative boosting
algorithm. Our attempts to do so revealed impractical.
Instead, we limited the use of a large CNN
to the {\it final\/} cascade stage. Upon learning the cascade, we
simply used a large CNN classifier as the final weak learner $g$
of~(\ref{eq:update}). Note that this has no
loss of optimality, since $\alpha$ was learned
with~(\ref{eq:a*}). The CNN is simply a descent direction of
(\ref{eq:bestLag}) unavailable to prior stages.
It differs from the standard proposal+CNN approach in that
1) not only the bounding boxes but also the confidence scores of the cascade
are forwarded to the deep CNN stage, and 2) the combination of the
proposal mechanism (cascade) and large CNN is optimal under the
well defined risk of~(\ref{equ:lagrangian}).

In our implementation, we considered both the
Alex \cite{DBLP:conf/nips/KrizhevskySH12} and
VGG \cite{DBLP:journals/corr/SimonyanZ14a} models. Previous
implementations \cite{DBLP:conf/cvpr/GirshickDDM14,
DBLP:journals/corr/HosangOBS15} have warped cropped patches to size
227$\times$227. However, such large patches are computationally expensive.
We adopted the convolutional layers from the pre-trained models and two
(randomly initialized) fully connected layers of 2048 units each.
These networks were fine tuned to the pedestrian datasets using
Caffe \cite{DBLP:conf/mm/JiaSDKLGGD14}. This
allowed us to use the canonical 128$\times$64 size for the pedestrian
template. For Alex-Net, we used a convolution stride of $2$ on the
first layer, instead of $4$ in the original model. For VGG-Net, we used
all aspects of the original configuration other than input size and fully
connected layers. While the original VGG-Net is approximately
8 times slower than the Alex-Net, the modified VGG-Net is only twice as
slow.

%%%%%%%%%%%%%%%%%%%%%%%%%%%% stage configuration %%%%%%%%%%%%%%%%%%%%%%%%%%%
\begin{figure}[!t]
\centering
\centerline{\epsfig{figure=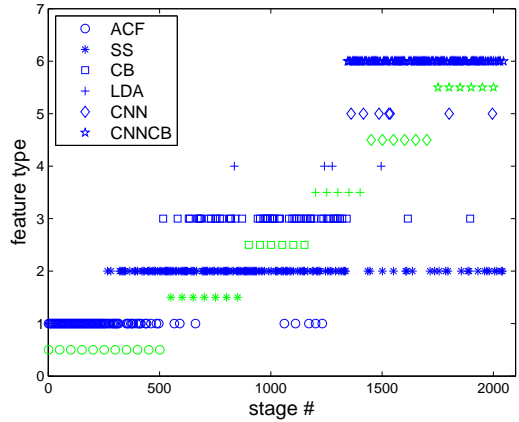,width=8cm,height=6cm}}
\caption{Stage configuration of the proposed CompACT cascade (blue) and
the manually set cascade (green). Only one in five (fifty) stages is shown
for the CompACT (manual) cascade.}
\label{fig:stages}
\end{figure}

%============================================================================================
\section{Experiments}

Various experiments were performed to evaluate the performance of CompACT
cascades. All times reported are for implementation on a
single CPU core (2.10GHz) of an Intel Xeon E5-2620 server with 64GB of
RAM. An NVIDIA Tesla K40M GPU was used for CNN computations.

%-------------------------------------------------------------------------
\subsection{Cascade Configuration}
\label{subsec:configuration}

We started by learning a CompACT cascade on the Caltech pedestrian dataset,
using the set up of \cite{DBLP:journals/pami/DollarABP14}.
The cascade used 2048 decision trees of depth $2$, and
was bootstrapped 6 times during training, after stages
$\{32,128,256,512,1024,1536\}$, using the procedure of
\cite{DBLP:conf/cvpr/GallL09,DBLP:conf/bmvc/TangLK12}.
Figure \ref{fig:stages} presents the
configuration of the learned cascade, showing how features of different
complexities were chosen at different stages. ACF features, which are the
cheapest, were the only selected for the first 200 stages,
and rarely chosen after stage 500. This suggests that the these features
are very efficient but not very discriminant. A better trade-off between
these two goals is achieved by the SS features,
which were selected throughout the training process. It is particularly
interesting that these features are competitive even for the later
cascade stages. This suggests that they can be very discriminant
despite their simplicity. Similarly, CB features were selected across
a large range of cascade stages. This is unlike LDA features,
which were rarely selected. These features do not appear to achieve a
good trade-off between discrimination and complexity. More surprisingly,
the CNN features were also rarely selected, with CNNCB dominating the late
cascade stages. This suggests that the CNNCB representation is more
discriminant. Recall that, while the CNN features are a little more efficient,
CompACT boosting weighs complexity less heavily than discrimination in
the late cascade stages.

%%%%%%%%%%%%%%%%%%%%%% TAB: cascade comparison %%%%%%%%%%%%%%%%%%%%%%%%%%%%%
\begin{table}[t]
\centering \scriptsize \setlength{\tabcolsep}{3.0pt}
\vspace{0.1cm} \caption{Comparison to single-feature cascades (MR: log-average miss-rate).}
\label{tab:cascade comparison single}
\begin{tabular}
{|c||cccccc|cc|}\hline
\multirow{2}{*}{Method}
&\multicolumn{6}{|c|}{Single Type}
&\multicolumn{2}{|c|}{CompACT}\\
\cline{2-9}
& ACF & SS & CB & LDA & CNN & CNNCB & ACF & CNN \\\hline
MR         &42.6  &34.29  &37.89 &37.15 &28.07 &26.93 &32.15 &23.82\\
time (s)   &0.07  &0.08   &0.23  &0.16  &0.87  &2.05  &0.11  &0.28\\\hline
\end{tabular}
\end{table}

%-------------------------------------------------------------------------
\subsection{Cascade Comparison}
\label{subsec:cascade comp}

The CompACT cascade of the previous section was compared to cascades
of other architectures. Table \ref{tab:cascade comparison single}
presents a comparison to the predominant architecture in the literature:
cascades of a single feature type. In this case, the
complexity penalty of~(\ref{equ:final lagragian}) is equal for all weak
learners, and CompACT reduces to standard boosting. This
was used to produce ``standard''
cascades of ACF, SS, CB, LDA, CNN and CNNCB features.
We start by noting that the implemented ACF outperforms \cite{DBLP:journals/pami/DollarABP14}. This is due to the use of a different
bootstrapping strategy.
Clearly, SS outperforms the other ACF-based features (ACF, CB, and LDA),
achieving higher accuracy {\it and\/} speed. This confirms
Figure~\ref{fig:stages}, where SS features were selected
throughout the detector. CB and LDA are more discriminant than ACF,
but have higher complexity. CNN features have higher accuracy than all
ACF-based features at the cost of a ten-fold increase in complexity over
ACF. Finally, CNNCB has the best detection results, but only a
marginal gain over CNN and much higher computation.
When compared to CompACT cascades, all single feature cascades
perform poorly. CompACT-ACF, which is restricted to ACF-based features,
has higher accuracy than all ACF-based single feature cascades
and is faster than most. CompACT-CNN, which
includes all features, has the best detection performance.
Note that not only its detection performance is clearly superior to the
best single-feature cascade (CNNCB) but it is also 10 times {\it faster\/}.

%%%%%%%%%%%%%%%%%%%%%% TAB: cascade comparison %%%%%%%%%%%%%%%%%%%%%%%%%%%%%
\begin{table}[t]
\centering \scriptsize \setlength{\tabcolsep}{3.0pt}
\caption{Comparison to multiple-feature cascades.}
\label{tab:cascade comparison multi}
\begin{tabular}
{|c||ccc|ccc|}\hline
\multirow{2}{*}{Method}
&\multicolumn{3}{|c|}{ACF-based}
&\multicolumn{3}{|c|}{ACF-based+Small CNN}\\
\cline{2-7}
& Boosting & Manual & CompACT & Boosting & Manual & CompACT \\\hline
MR     &33.06 &36.08 &32.15 &22.37 &25.46 &23.82\\
time (s) &0.41  &0.11  &0.11  &2.69  &0.28  &0.28\\\hline
\end{tabular}
\end{table}

Table \ref{tab:cascade comparison multi} presents a comparison to cascades
that combine multiple features. ``Boosting''  is a cascade learned without
complexity constraints ($\eta = 0$ in (\ref{equ:final lagragian})). This is equivalent to applying existing cascade learning
algorithms to the diverse feature set considered in this work.
``Manual'' is an attempt to ``hand-code'' the behavior of CompACT,
by restricting the boosting algorithm without complexity
constraint ($\eta = 0$)  to use certain types of features in different
cascade stages. This restriction is based on feature complexity, as
illustrated in Figure~\ref{fig:stages}. The features were ranked by complexity
and used sequentially, each feature type being used in approximately 400
stages. The two sides of Table \ref{tab:cascade comparison multi} differ in
that only ACF-based features were used on the left, while both these
and the small CNN model were used on the right.
In both cases, the ``manual'' cascade has low complexity but
poor accuracy. ``Boosting,'' on the other hand, can
produce a more accurate cascade. The price is, however, a significant
increase in complexity. CompACT achieves the best trade-off between
accuracy and complexity. Note also the introduction of the small CNN model enables substantially better
cascades, as long as a complexity penalty is assigned to it
during learning.

%%%%%%%%%%%%%%%%%%%%%% TAB: CNN comparison %%%%%%%%%%%%%%%%%%%%%%%%%%%%%
\begin{table}[t]
\centering \scriptsize \setlength{\tabcolsep}{3.0pt}
\caption{Performance of CompACT cascades using large CNNs.}
\label{tab:CNN comparison}
\begin{tabular}
{|c||c||cc|cc|cc|}\hline
\multirow{2}{*}{Method}
&\multirow{2}{*}{CompACT}
&\multicolumn{2}{|c|}{Proposals}
&\multicolumn{2}{|c|}{Intermediate}
&\multicolumn{2}{|c|}{Embedded}\\
\cline{3-8}
&\multicolumn{1}{|c||}{} & Alex & VGG & Alex & VGG & Alex & VGG \\\hline
MR       &18.92  &19.59   &14.77   &16.18   &13.71  &14.96  &11.75\\
time (s)   &0.25   &+0.01   &+0.03   &+0.01   &+0.03  &+0.1   &+0.25\\\hline
\end{tabular}
\end{table}

%-------------------------------------------------------------------------
\subsection{Large CNN models}
\label{subsec:deepcacas}

While the previous experiments only use small models, a number of
experiments were performed with large models. These experiments
were performed on both Caltech and KITTI, in both cases
using cascades of 4096 decision trees of depth 5. These were bootstrapped
9 times, after stages $\{32,128,256,512,1024,1536,2048,2560,3328\}$. For Caltech, we used the training set size of \cite{DBLP:conf/nips/NamDH14}, and the template size 64$\times$32 as in \cite{DBLP:journals/pami/DollarABP14}. On KITTI, test images were upsampled by 2 to detect pedestrians of
height $25$. This enabled the use of a single
template size. After upsampling, the detected bounding boxes
(minimum height of 50) had twice the actual object size. They were
rescaled down by a factor of 2.

Table \ref{tab:CNN comparison} compares the performance of the CompACT
cascade with small CNNs (denoted CompACT) with several variants for
the inclusion of large CNNs. In all these variants, the large CNN is
computed only on windows selected by CompACT. The times noted as "+"
reflect the added cost of running the image patches through it.
The ``Proposal'' columns report to the use of the CompACT cascade as a
proposal mechanism~\cite{DBLP:conf/cvpr/GirshickDDM14,
DBLP:journals/corr/HosangOBS15} for the CNN. The ``Embedded'' columns
report to the use of the large CNN as the last stage
of the cascade, as discussed in Section~\ref{subsec:big cnn}. Finally, the
``Intermediate'' columns report to an intermediate
between these two architectures. As with proposals,
the large CNN stage was only applied to the CompACT output, after
non-maximum suppression (NMS). However, the prediction was that
of~(\ref{eq:update}), i.e. the CNN and CompACT scores were {\it combined,\/}
 using the coefficient $\alpha$ learned by boosting.

%%%%%%%%%%%%%%%%%%%%%%%%%%%% Caltech Pedestrian Detection %%%%%%%%%%%%%%%%%%%%%%%%%%%
\begin{figure}[!t]
\centering
\centerline{\epsfig{figure=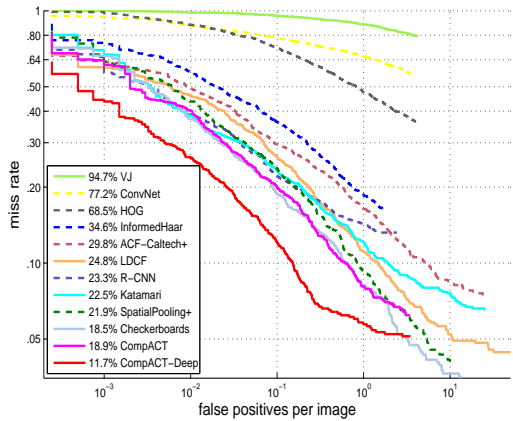,width=8cm,height=6cm}}
\caption{Comparison to state-of-the-art on Caltech (reasonable).}
\label{fig:caltech}
\end{figure}

A number of interesting conclusions are possible. First,
under the proposal architecture, only VGG improved on the
CompACT cascade. For Alex, there was no
benefit. This shows that the CompACT cascade is already a very good
classifier. Second, the {\it embedding\/} of the large CNN on the
CompACT model achieved the best results in all cases. This shows
that the ComPACT cascade score contains information that complements that
of the CNN scores. For both CNN models, it was better to combine scores
with the CompACT cascade than to consider the latter simply as a proposal
mechanism. Finally, the theoretically
more sound embedding of the large CNN before NMS (''Embedding'') always produced higher detection accuracy than the combination after NMS (``Intermediate'').
This, however, had substantially less computation, since the number of bounding
boxes is approximately 10 times smaller after NMS.

%-------------------------------------------------------------------------
\subsection{Comparison with the state-of-the-art}
\label{subsec:pedestrian comp}

Figure \ref{fig:caltech} compares two CompACT pedestrian detectors
to the state of the art on Caltech.
CompACT refers to the model using ``ACF + small CNN features'', and
CompACT-Deep to the model with the embedded VGG model in the last stage.
CompACT  achieves state-of-the-art performance, close
to \cite{DBLP:journals/corr/ZhangBS15}. Note that the competing detectors
- Katamari \cite{Benenson2014Eccvw} and
SpatialPooling+ \cite{DBLP:conf/eccv/PaisitkriangkraiSH14} - combine
many features (HOG, LBP, spatial covariance, optical flow, multiple
detectors, etc.) and are all quite slow. The same holds for the
state-of-the-art implementation of Checkerboards, which
requires a large number of filter
channels \cite{DBLP:journals/corr/ZhangBS15}. On the other hand, CompACT
runs at 4 fps on a relatively slow processor. The CompACT-Deep cascade
performs even better - 7 points better than the
state-of-the-art \cite{DBLP:journals/corr/ZhangBS15} and 11 points better
than the best deep pedestrian detector \cite{DBLP:journals/corr/HosangOBS15}!
CompACT-Deep runs at 2fps and is faster than the competing
detectors \cite{Benenson2014Eccvw,DBLP:conf/eccv/PaisitkriangkraiSH14,
DBLP:journals/corr/ZhangBS15}.

%%%%%%%%%%%%%%%%%%%%%%%%% KITTI Pedestrian Detection %%%%%%%%%%%%%%%%%%%%%%%%
\begin{figure}[!t]
\centering
\centerline{\epsfig{figure=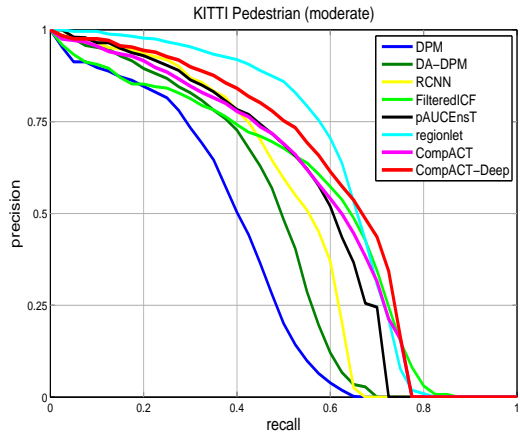,width=8cm,height=6cm}}
\caption{Comparison to state-of-the-art on KITTI Pedestrian (moderate).}
\label{fig:kitti}
\end{figure}

Figure \ref{fig:kitti} and Table \ref{tab:kitti AUC and Time} summarize
performance on KITTI. Since test images are larger than in Caltech,
running times are higher on this dataset. Nevertheless, the CompACT cascade
is the fastest of all the state-of-the-art detectors. Note that it uses
approximately the same number of feature channels (including the CNN model)
as pAUCEnsT \cite{DBLP:conf/eccv/PaisitkriangkraiSH14} and
FilteredICF \cite{DBLP:journals/corr/ZhangBS15}, which are both much
less accurate and slower. R-CNN \cite{DBLP:journals/corr/HosangOBS15,
DBLP:conf/cvpr/GirshickDDM14}, the only CNN detector on KITTI,
is also substantially weaker than CompACT-Deep (difference larger than 8
points). Overall, the only approach competitive with the CompACT-Deep cascade
is the Regionlets method of \cite{DBLP:conf/iccv/WangYZL13}. However,
this work only reports classification times, excluding the time needed
to generate proposals, which can be on the order of several seconds.
This is equivalent to only accounting for the processing time of
the last stage of the CompACT-Deep model, which is 0.25 second.

%%%%%%%%%%%%%%%%%%%%%% KITTI Detailed comparison %%%%%%%%%%%%%%%%%%%%%%%%%%%%%
\begin{table}[t]
\centering \scriptsize \setlength{\tabcolsep}{3.0pt}
\vspace{0.1cm} \caption{Comparison to state-of-the-art detectors on KITTI.
Note: $^*$ ignores the time needed to compute object proposals.}
\label{tab:kitti AUC and Time}
\begin{tabular}
{|c||ccc||c|}\hline\multicolumn{1}{|c||}{Methods}
&\multicolumn{1}{c}{Easy}
&\multicolumn{1}{c}{Moderate}
&\multicolumn{1}{c||}{Hard}
&\multicolumn{1}{c|}{Time (s)}\\\hline
DPM             &45.50  &38.35  &34.78  &10 \\
DA-DPM          &56.36  &45.51  &41.08  &21 \\
RCNN            &61.61  &50.13  &44.79  &4 \\
FilteredICF     &61.14  &53.98  &49.29  &40 \\
pAUCEnsT        &65.26  &54.49  &48.60  &60 \\
regionlet       &73.14  &61.15  &55.21  &$1^*$ \\
CompACT         &65.35  &54.92  &49.23  &0.75 \\
CompACT-Deep    &70.69  &58.74  &52.71  &1 \\\hline
\end{tabular}
\end{table}

%============================================================================================
\section{Conclusion}

In this work, we proposed the CompACT boosting algorithm for learning
complexity-aware detector cascades. By optimizing classification risk under
a complexity constraint, CompACT produces cascades that push features of
high complexity to the later cascade stages. This has been shown to enable
the seamless integration of multiple feature families in a unified design. This integration extends to
features, such as deep CNNs, that were previously
beyond the realm of cascaded detectors. The proposed CompACT cascades
also generalize the popular combination of object
proposals+CNN, which they were shown to outperform. Finally, we have
shown that a pedestrian detector learned by application of CompACT to a
diverse feature pool achieves state-of-the-art detection rates on
Caltech and KITTI, with much faster speeds than competing methods.

% In this paper, we propose a complexity-aware cascade for pedestrian detection, in which the classification risk is optimized under a constraint of the complexity risk. Multiple types of features covering a wide range of complexities are used, and the learned cascade interestingly push the cheap features to the beginning stages and the expensive ones to the later stages. This property makes the cascade own a good accuracy vs. complexity trade-off. The experiments on two public pedestrian detection datasets show the proposed detector has achieved the state of the art performance while running faster than most of the closest competing detectors.

{\small
\bibliographystyle{ieee}
\bibliography{egbib}
}

\end{document}